\documentclass[runningheads]{llncs}
\usepackage{graphicx}
\usepackage{color}
\usepackage{hyperref}
\usepackage{kbordermatrix}
\usepackage{amsmath}

\begin{document}

\title{NLNDE: The Neither-Language-Nor-Domain-Experts' Way of Spanish Medical Document De-Identification}
\subtitle{}

\titlerunning{The NLNDE Way of Spanish Medical Document De-Identification }

\author{Lukas Lange\inst{1,2,3}
	\and Heike Adel\inst{1}
	\and Jannik Str\"otgen\inst{1}
}

\authorrunning{L. Lange et al.}

\institute{	
	Bosch Center for Artificial Intelligence \\Robert-Bosch-Campus 1, 71272 Renningen, Germany \\
	\email{\{Lukas.Lange,Heike.Adel,Jannik.Stroetgen\}@de.bosch.com} \\
	\url{https://www.bosch-ai.com}
	\and Spoken Language Systems (LSV),
	\and Saarbr\"ucken Graduate School of Computer Science \\
	Saarland Informatics Campus, Saarland University, Saarbr\"ucken, Germany
}

\maketitle

\begin{abstract}
	Natural language processing has huge potential in the medical domain which recently led to a lot of research in this field. 
	However, a prerequisite of secure processing of medical documents, 
	e.g., patient notes and clinical trials, is the proper de-identification of privacy-sensitive information.   
	In this paper, we describe our NLNDE system, with which we participated in the MEDDOCAN competition, 
	the medical document anonymization task of IberLEF 2019. We address the task of detecting and 
	classifying protected health information from Spanish data as a sequence-labeling problem 
	and investigate different embedding methods for our neural network.
	Despite dealing in a non-standard language and domain setting, the NLNDE system achieves promising results in the competition.
\end{abstract}

\section{Introduction}\label{sec:introduction}

The anonymization of privacy-sensitive information is of increasing importance
in the age of digitalization and machine learning.
It is, in particular, relevant for texts from the medical domain
that contain a large number of sensitive information by nature.
The shared task MEDDOCAN (Medical Document Anonymization) \cite{meddocan2019} aims at
automatically detecting protected health information (PHI) from Spanish
medical documents.
Following the past de-identification task on English PubMed abstracts \cite{i2b2/Stubs15},
it is the first competition on this topic on Spanish data.

In this paper, we describe our submissions to MEDDOCAN and their results.
We, as \textbf{N}either \textbf{L}anguage \textbf{N}or \textbf{D}omain \textbf{E}xperts (NLNDE),
address the anonymization task as a sequence-labeling problem and 
use a combination of different state-of-the-art approaches from natural language processing
to tackle its challenges.

We train recurrent neural networks with conditional random field output layers
which are state of the art for different sequence labeling tasks, such
as named entity recognition \cite{ner/Lample16}, part-of-speech tagging \cite{kemos2019}
or de-identification \cite{dei/Khin18,dei/Liu17}.
Recently, the field of natural language processing has seen another
boost in performance by using context-aware language representations which
are pre-trained on a large amount of unlabeled corpora \cite{flair/Akbik19,bert/Devlin19,elmo/Peters18}.
Therefore, we experiment with FLAIR embeddings for Spanish \cite{flair/Akbik19}
to represent the input of our networks. 
In our different runs, we further explore the advantages of domain-specific fastText embeddings \cite{fastText/bojanowski2017}
that have been pre-trained on SciELO and Wikipedia articles \cite{emb/Soares19}.

From a natural-language-processing perspective, the MEDDOCAN task
is interesting due to the non-standard domain (medicine) and language (Spanish) of the documents.
The results of our submissions show that state-of-the-art architectures
for sequence-labeling tasks can be directly transferred to these settings
and that domain-specific embeddings are helpful but not necessary.
\section{Methods}\label{sec:approach}
In this section, we first give an overview of the different embedding methods we use in our system.
Second, we describe the architecture of our system.

\begin{figure}
	\centering
	\includegraphics[width=\textwidth]{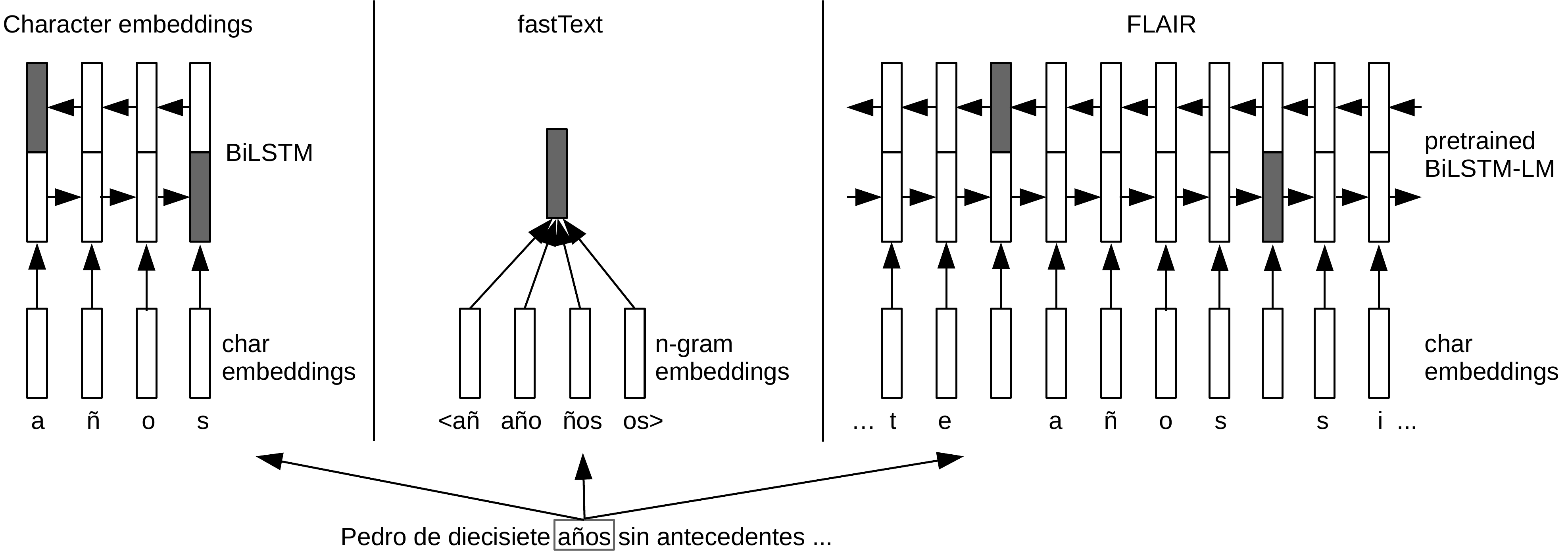}
	\caption{Comparison of sub-word embedding methods. The shaded vectors are used to represent the input token.}
	\label{fig:embeddings}
\end{figure}

\subsection{Sub-word Embedding Methods} \label{sec:embeddings}
In our different runs, we investigate the impact of the following sub-word embedding methods:
character-based, fastText and FLAIR embeddings. They are depicted in Figure \ref{fig:embeddings}.

\paragraph{Character Embedding:} 
The characters of a word are represented by randomly initialized embeddings. Those are passed to a bi-directional long short-term memory network (BiLSTM).
The last hidden states of the forward and backward pass are concatenated to represent the word~\cite{ner/Lample16}.

\paragraph{FastText Embedding:} The fastText embeddings represent a word by the normalized sum of the embeddings for the n-grams of the word~\cite{fastText/bojanowski2017}.

\paragraph{FLAIR Embedding:} FLAIR computes character-based embeddings for each word depending on all words in the context~\cite{flair/Akbik18}. 
For this, the complete sentence is used as the input to the BiLSTM instead of only a single word. The BiLSTM of FLAIR is pretrained 
using a character-level language model objective, i.e., given a sequence of characters, compute the probability for the next character.

\subsection{NLNDE System} \label{sec:system}
\begin{figure}
	\centering
	\includegraphics[width=.8\textwidth]{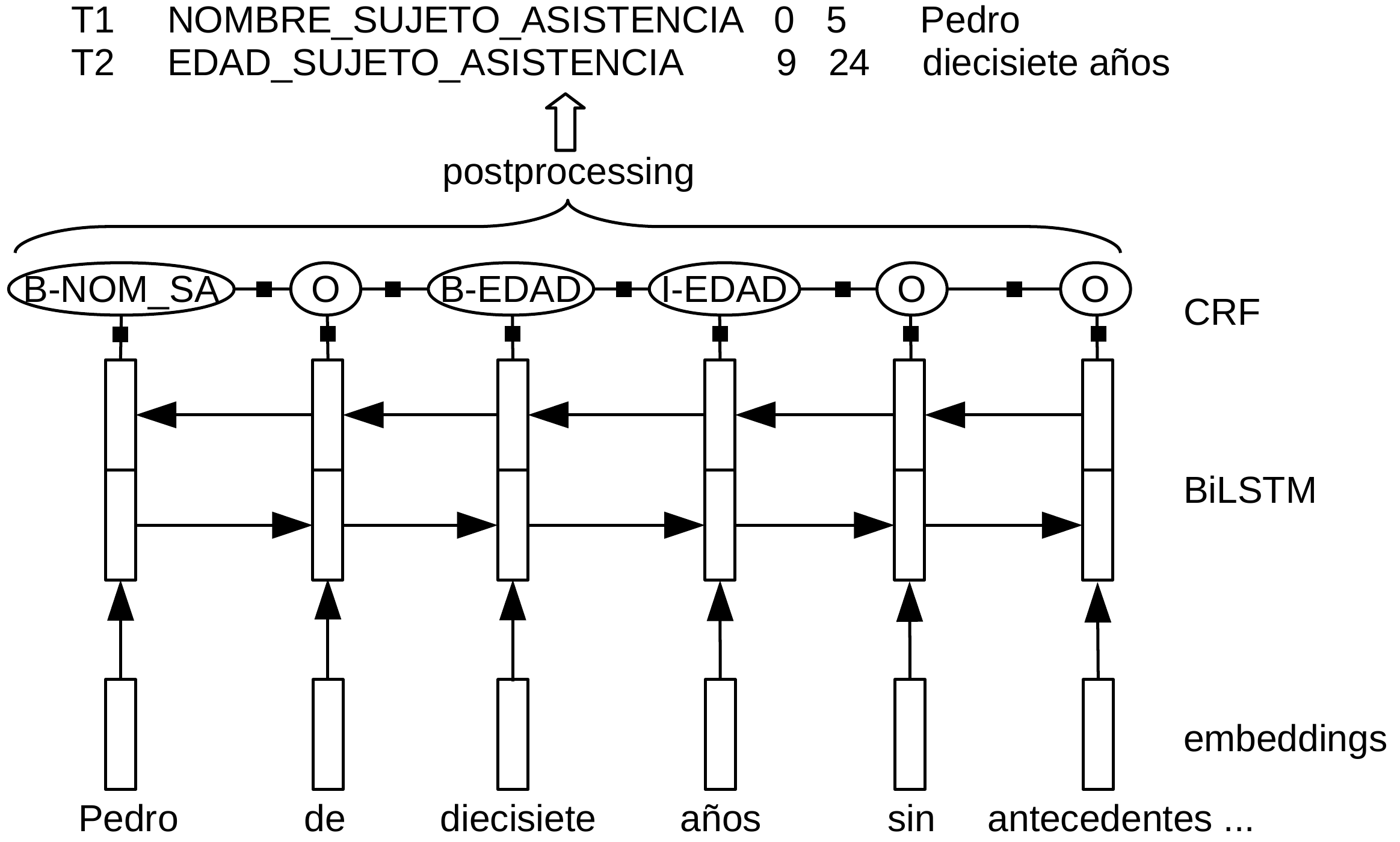}
	\caption{General architecture of all our models. The label prefixes ``B-'' and ``I-'' show how we address the task as a sequence-labeling task.}
	\label{fig:model}
\end{figure}

In Figure~\ref{fig:model}, the architecture of our model is depicted. 
In the following, we explain the different layers.

\paragraph{Input Representation.}

We tokenize the input using the tokenizer provided by the shared task organizers~\cite{spaccc/Intxaurrondo19}.
Then, we represent each token with embeddings. In our runs, we investigate the impact of the following kinds of embeddings: 
the output of an LSTM over character embeddings (50 dimensions, randomly initialized and fine-tuned
during training), 
domain-independent fastText embeddings (300 dimensions, pre-trained on Spanish text \cite{fastText/Grave18}),
domain-specific fastText embeddings (100 dimensions, pre-trained on Spanish SciELO and Wikipedia articles \cite{emb/Soares19}),
and FLAIR embeddings (4096 dimensions, pre-trained on Spanish text \cite{flair/Akbik18}). 
For FLAIR embeddings, we also test their pooled version (8192 dimensions, using min pooling) \cite{flair/Akbik19}.
Note that except for the character embeddings, we do not fine-tune any of the embeddings.

\paragraph{BiLSTM-CRF Layers.}

The embeddings are fed into a BiLSTM with a conditional 
random field (CRF) output layer, similar as done by Lample et al. \cite{ner/Lample16}. 
The CRF output layer is a linear-chain CRF, i.e., it learns transition scores
between the output classes. For training, the forward algorithm
is used to sum the scores for all possible sequences. During decoding, the Viterbi algorithm is applied to
obtain the sequence with the maximum score.
Note that the hyperparameters are the same across all runs. 
We use a BiLSTM hidden size of 256
and train the network with mini-batch stochastic gradient descent using a learning rate of 0.1 and a batch size of 32. 
For regularization, we employ early stopping on the development set and
apply dropout with probability 0.5 on the input representations.

\paragraph{Postprocessing.}

The output of the model is further adjusted with a post-process\-ing layer, similar as done by Yang et al. \cite{ib2b/Yang14} and Liu et al. \cite{dei/Liu17}.
As some classes from the annotation guidelines~\footnote{\url{http://temu.bsc.es/meddocan/index.php/annotation-guidelines/}} do not occur in the training data,
we tackle them with pattern matching. For this, we use regular expressions
for URLs, IP- and MAC addresses
to detect the classes \texttt{URL\_WEB} and \texttt{DIREC\_PROT\_INTERNET}, overwriting the results of the neural classifier. 

\section{Submissions}\label{sec:submissions}

We submitted five runs to the MEDDOCAN competition.
All of them are based on the architecture described in Section~\ref{sec:system}. 
They only differ in the usage of different input representations.

\begin{itemize}
	\item[S1] (\textit{Char+fastText+Domain}):  Our first run uses a combination of
	character embeddings, domain-independent fastText 
	embeddings as well as domain-specific fastText embeddings to represent tokens. The resulting representation for each token has 450 dimensions. 
	
	\item[S2] (\textit{FLAIR+fastText}): In contrast to all other runs, the second run uses only domain-independent embeddings, 
	i.e., embeddings that have been trained on standard narrative and news data from Common Crawl and Wikipedia. 
	In particular, it uses a combination of domain-independent fastText embeddings and Flair embeddings. 
	
	\item[S3] (\textit{FLAIR+fastText+Domain}): The third run adds domain-specific fastText 
	embeddings to the system of the second run in order to investigate the impact of domain knowledge. 
	
	\item[S4] (\textit{PooledFLAIR}): The fourth run is equal to the third run, except that we use the minimum-pooling version of the FLAIR embeddings.
	
	\item[S5] (\textit{Ensemble}): The fifth run is an ensemble of the previous four runs using weighted voting: 
	Each classifier $C_i$ is assigned a weight $w_i \in [0.5, 3]$. For each output label, the weights of the classifiers predicting it are summed.
	Then, the label with the highest score is chosen if it exceeds a specific threshold $t \in [1,5]$, or O (no PHI class) otherwise.
	The weights    and threshold are selected based on results on the development set as follows: $w_1 = 0.5$, $w_2 = 2.0$, $w_3 = 2.5$, $w_4 = 0.5$ and $t = 3$. 
	With these parameters, a label needs votes from at least two classifiers ($w_i < t, i \in \left\{1,2,3,4\right\}$). 
	However, the models of the submissions S2 and S3 
	are assigned higher weights than S1 and S4. This reflects their performance (see next section). 
\end{itemize}
\section{Results and Analysis}\label{sec:results}

This section describes our results and analysis.
We report the results on the MEDDOCAN test set using the official shared task evaluation measures \cite{meddocan2019}. 

\subsection{Results for Task 1: NER Offset and Entity Type Classification}
In the first sub-task, the systems need to find spans for de-identification and categorize them into one of 29 classes.
Table \ref{tab:task1} presents our results on this sub-task.

\begin{table}
	\centering
	\caption{Results of our five runs for Task 1.}
	\begin{tabular}{l|cccc}
		\textbf{S$_{ID}$} & \textbf{Leak} & \textbf{Precision} & \textbf{Recall} & \textbf{F1} \\ \hline
		S1 (\textit{Char+fastText+Domain}) & 0.02432 & 0.96956 & 0.96767 & 0.96861 \\ 
		S2 (\textit{FLAIR+fastText}) & 0.02378 & \textbf{0.97078} & 0.96838 & 0.96958 \\ 
		S3 (\textit{FLAIR+fastText+Domain}) & \textbf{0.02299} & 0.96978 & \textbf{0.96944} & \textbf{0.96961} \\ 
		S4 (\textit{PooledFLAIR}) & 0.02724 & 0.96720  & 0.96379 & 0.96549 \\
		S5 (\textit{Ensemble}) & 0.02365 & 0.97044 & 0.96856 & 0.96950  \\ 
	\end{tabular}
	\label{tab:task1}
\end{table}

While the domain-independent system (run 2 with FLAIR and domain-independent fastText embeddings) leads to the
highest recall values, the third run that also uses domain-specific fastText embeddings
achieves the highest F1 scores. This shows that integrating domain knowledge into the token representation is beneficial.
However, the differences among the five runs are rather small, indicating that the architecture itself is
already strong enough for the given dataset and the impact of different input representations is minor.

\subsection{Results for Task 2: Sensitive Token Detection}
Tables \ref{tab:task2a} and \ref{tab:task2b} provide the results of our models on the second sub-task
(sensitive token detection). In contrast to task 1, this is a binary classification task. 
Since the official evaluation measure for this task is the strict one,
we focus our explanation on Table \ref{tab:task2a}. 
The main ranking of our models is the same as the ranking for sub-task 1:
the addition of domain-specific input representations performs best. Interestingly, the domain-specific
input representations (run 3) now perform best in terms of recall as well while the domain-independent input
representations (run 2) perform best in terms of precision.

\begin{table}
	\centering
	\caption{Results of our five runs for Task 2 (Evaluation: Strict).}
	\begin{tabular}{l|ccc}
		\textbf{S$_{ID}$} & \textbf{Precision} & \textbf{Recall} & \textbf{F1} \\ \hline
		S1 (\textit{Char+fastText+Domain}) & 0.97522 & 0.97333 & 0.97427 \\
		S2 (\textit{FLAIR+fastText}) & \textbf{0.97574} & 0.97333 & 0.97453 \\
		S3 (\textit{FLAIR+fastText+Domain}) & 0.97508 & \textbf{0.97474} & \textbf{0.97491} \\
		S4 (\textit{PooledFLAIR}) & 0.97217 & 0.96873 & 0.97045 \\
		S5 (\textit{Ensemble}) & 0.97540  & 0.97350  & 0.97445 \\ 
	\end{tabular}
	\label{tab:task2a}
\end{table}

\begin{table}
	\vspace{-10mm}
	\centering
	\caption{Results of our five runs for Task 2 (Evaluation: Merged).}
	\begin{tabular}{l|ccc}
		\textbf{S$_{ID}$} & \textbf{Precision} & \textbf{Recall} & \textbf{F1} \\ \hline
		S1 (\textit{Char+fastText+Domain}) & \textbf{0.98749} & \textbf{0.98311} & \textbf{0.9853} \\
		S2 (\textit{FLAIR+fastText}) & 0.98648 & 0.98145 & 0.98396 \\
		S3 (\textit{FLAIR+fastText+Domain}) & 0.98566 & 0.98264 & 0.98415 \\
		S4 (\textit{PooledFLAIR}) & 0.98182 & 0.97730  & 0.97956 \\
		S5 (\textit{Ensemble}) & 0.98598 & 0.98162 & 0.98380     \\
	\end{tabular}
	\label{tab:task2b}
	\vspace{-5mm}
\end{table}

In both sub-tasks, FLAIR embeddings outperform standard character embeddings (except for the evaluation
type merge in Table \ref{tab:task2b}). Also, for both sub-tasks, pooling of FLAIR embeddings leads to worse results.
Surprisingly, run 5, i.e., the ensemble of the models from runs 1--4, does not improve the results over single models.

\newcommand{\rb}[1]{\raisebox{3ex}{\rotatebox[origin=c]{90}{#1}}}

\begin{table}[t]
	\centering
	\hspace{-5mm}
	\caption{Confusion matrix of the best model (S3) on the development set.\protect\footnote[2]}
	\begin{tabular}{c@{}c@{}}

		& \mbox{\hspace{6mm}Predicted Label} \\ 
		
		\parbox[c][20mm][t]{4mm}{\rotatebox{90}{Gold Label}} &
		\tiny
		
		$ \kbordermatrix{
			
			& {\rb{O}} 
			& {\rb{CALLE}} 
			& {\rb{CS}} 
			& {\rb{MAIL}} 
			& {\rb{EDAD}} 
			& {\rb{FAM}} 
			& {\rb{FECHA}} 
			& {\rb{HOS}} 
			& {\rb{ID\_AS}} 
			& {\rb{ID\_CON}} 
			& {\rb{ID\_EPS}} 
			& {\rb{ID\_SUJ}} 
			& {\rb{ID\_TPS}} 
			& {\rb{INST}} 
			& {\rb{NOM\_PS}} 
			& {\rb{NOM\_SA}} 
			& {\rb{\#FAX}} 
			& {\rb{\#TEL}} 
			& {\rb{OTRO}} 
			& {\rb{PAIS}} 
			& {\rb{PROF}} 
			& {\rb{SEXO}} 
			& {\rb{TER}} \\
			
			O & 123293 & 21 & & & 2 & 23 & 9 & 3 & & 1 & & 4 & 6 & 28 & 3 & & 1 & & & 2 & & 3 & \\
			CALLE & 15 & 2997 & & & & & & 2 & & & & & & 1 & 8 & 1 & & & & & & & 6 \\
			CS & & & 8 & & & & & 3 & & & & & & & & & & & & & & & \\
			MAIL & & & & 256 & & & & & & & & & & & 3 & & & & & & & & \\
			EDAD & 14 & & & & 1014 & 3 & & & & & & & & & & & & & & & & & \\
			FAM & 18 & & & & 2 & 104 & & & & & & 2 & & & & & & & 2 & & & & \\
			FECHA & 16 & & & & & & 1089 & & & & & & & & & & & & & & & & \\
			HOS & 10 & 4 & & & & & 4 & 551 & & & & & & 11 & & & & & & & & & 1 \\
			ID\_AS & & & & & & & & & 573 & & & 2 & 6 & & & & & & & & & & \\
			ID\_CON & & & & & & & & & & 32 & & & & & & & & & & & & & \\
			ID\_EPS & 4 & & & & & & & & & & & & & & & & & & & & & & \\
			ID\_SA & 9 & & & & & & & & & & & 293 & & & & & & & & & 2 & & \\
			ID\_TPS & & & & & & & & & & & & & 663 & & & & & & & & & 1 & \\
			INST & 35 & 8 & 3 & & & & & 11 & & & & & & 190 & & & & & & & & & \\
			NOM\_PS & 3 & & & & & 1 & & & & & & & & & 1585 & 2 & & & & & & & \\
			NOM\_SA & 3 & & & & & & & & & & & & & & & 779 & & & & & & & \\
			\#FAX & & & & & & & & & & & & & & & & & 16 & & & & & & \\
			\#TEL & 1 & & & & & & & & & & & & & & & & & 70 & & & & & 1 \\
			OTRO & 11 & & & & & 2 & & & & & & 1 & & & & & & & 2 & & & & \\
			PAIS & & & & & & & & & & & & & & & & & & & & 349 & & & \\
			PROF & 8 & & & & & & & & & & & & & & & & & & & & 1 & & \\
			SEXO & & & & & & & & & & & & & & & & & & & & & & 456 & \\
			TER & 9 & 20 & & & & & 1 & 5 & & & & & & 5 & & & & & & 3 & & & 1141 \\
		} $ \\ 
	\end{tabular}
	\label{fig:confusion}
\end{table}

\footnotetext[2]{
	Abbreviations for entity types: \\ \tiny
	CALLE~(CALLE), 
	CENTRO\_SALUD~(CS), 
	CORREO\_ELECTRONICO~(MAIL), 
	EDAD\_SUJETO\_ASISTENCIA~(EDAD), 
	FAMILIARES\_SUJETO\_ASISTENCIA~(FAM),
	FECHAS~(FECHA), 
	HOSPITAL~(HOS), 
	ID\_ASEGURAMIENTO~(ID\_AS), 
	ID\_CONTACTO\_ASISTENCIAL~(ID\_CON), 
	ID\_EMPLEO\_PERSONAL\_SANITARIO~(ID\_EPS), 
	ID\_SUJETO\_ASISTENCIA~(ID\_SA), 
	ID\_TITULACION\_PERSONAL\_SANITARIO~(ID\_TPS), 
	INSTITUCION~(INST), 
	NOMBRE\_PERSONAL\_SANITARIO~(NOM\_PS),  
	NOMBRE\_SUJETO\_ASISTENCIA~(NOM\_SA),  
	NUMERO\_FAX~(\#FAX),  
	NUMERO\_TELEFONO~(\#TEL),  
	OTROS\_SUJETO\_ASISTENCIA~(OTRO),  
	PAIS~(PAIS),  
	PROFESION~(PROF),  
	SEXO\_SUJETO\_ASISTENCIA~(SEXO),  
	TERRITORIO~(TER)
}

\subsection{Confusion Matrix Analysis}
Table~\ref{fig:confusion} shows the confusion matrix of our best performing system (run 3). It is similar to the identity matrix, i.e., confusions between classes happen very rarely.
The most confusions happen with O, the label we assign to all non-PHI terms which might be caused by the high number of occurrences of this class in the training dataset. 
Confusions among PHI-classes happen mostly between related classes. For example, Hospital (HOS) and Institution (INST) are confused quite often, as Hospital is a subclass of Institution and other medical institutions are tagged with Hospital and vice versa, e.g., \textit{Clinica Gnation} is an institution tagged as a hospital. Analogously, Streets (CALLE) and Territoriums (TER) are getting confused often, as both classes are related and typically constitute of multiple tokens. In contrast to this, Countries (PAIS) are tagged correctly almost every time, as there is only a very limited number of countries and they are usually single token expressions. 

\vspace{-2mm}
\subsection{Synthetic Augmentation Case Study}
As mentioned above, the performance difference between our systems is rather small.
This may be caused by the synthetic augmentation of the MEDDOCAN data which was used to extend the texts with header and footer information containing many PHI terms. In fact, 85\% of PHI terms appear in the augmented text parts. While this extension is necessary to cover more classes and PHI terms, the synthetic nature of these extensions may have an impact on the performance of automatic classifiers. 
Therefore, we perform a case study in which we remove these parts from the test set and compare only the predictions found in the real text. Only 838 out of 5661 (14.8\%) annotations 
and only 13 out of 29 classes remain in this experiment. 
The performances of our systems are decreased to F1 scores around 0.90 which is still rather high.
This shows that our systems have learned more than just to reproduce the synthetic data augmentation.
However, the performance differences among our systems are still small, indicating that the data augmentation was not the reason for this behavior. 
Note, however, that we did not retrain our models without the synthetic augmentation.
\section{Conclusions}\label{sec:conclusion}
In this paper, we described the system with which we participated in the MEDDOCAN competition
on automatically detecting protected health information from Spanish
medical documents. As neither language nor domain experts, we addressed the task
with a sequence labeling model. In particular, we trained a bi-directional
long short-term memory network and explored different input representations.
All of our runs achieved high performance with F1 scores about 97\%.

\bibliographystyle{splncs04}
\bibliography{2019-IberLEF}

\begin{thebibliography}{10}
\providecommand{\url}[1]{\texttt{#1}}
\providecommand{\urlprefix}{URL }
\providecommand{\doi}[1]{https://doi.org/#1}

\bibitem{flair/Akbik19}
Akbik, A., Bergmann, T., Vollgraf, R.: Pooled contextualized embeddings for
  named entity recognition. In: Proc. of NAACL. pp. 724--728 (2019)

\bibitem{flair/Akbik18}
Akbik, A., Blythe, D., Vollgraf, R.: Contextual string embeddings for sequence
  labeling. In: Proc. of COLING. pp. 1638--1649 (2018)

\bibitem{fastText/bojanowski2017}
Bojanowski, P., Grave, E., Joulin, A., Mikolov, T.: Enriching word vectors with
  subword information. Transactions of the Association for Computational
  Linguistics  \textbf{5},  135--146 (2017). \doi{10.1162/tacl\_a\_00051}

\bibitem{bert/Devlin19}
Devlin, J., Chang, M.W., Lee, K., Toutanova, K.: {BERT}: Pre-training of deep
  bidirectional transformers for language understanding. In: Proc. of NAACL.
  pp. 4171--4186 (2019)

\bibitem{fastText/Grave18}
Grave, E., Bojanowski, P., Gupta, P., Joulin, A., Mikolov, T.: Learning word
  vectors for 157 languages. In: Proc. of LREC (2018)

\bibitem{spaccc/Intxaurrondo19}
Intxaurrondo, A.: {SPACCC} (spanish clinical case corpus) tokenizer (Mar 2019).
  \doi{10.5281/zenodo.2586978}

\bibitem{kemos2019}
Kemos, A., Adel, H., Sch{\"u}tze, H.: Neural semi-{M}arkov conditional random
  fields for robust character-based part-of-speech tagging. In: Proc. of NAACL.
  pp. 2736--2743 (2019)

\bibitem{dei/Khin18}
Khin, K., Burckhardt, P., Padman, R.: A deep learning architecture for
  de-identification of patient notes: Implementation and evaluation. CoRR
  \textbf{abs/1810.01570} (2018)

\bibitem{ner/Lample16}
Lample, G., Ballesteros, M., Subramanian, S., Kawakami, K., Dyer, C.: Neural
  architectures for named entity recognition. In: Proc. of NAACL (2016)

\bibitem{dei/Liu17}
Liu, Z., Tang, B., Wang, X., Chen, Q.: De-identification of clinical notes via
  recurrent neural network and conditional random field. Journal of Biomedical
  Informatics  \textbf{75},  S34 -- S42 (2017).
  \doi{https://doi.org/10.1016/j.jbi.2017.05.023}

\bibitem{meddocan2019}
Marimon, M., Gonzalez-Agirre, A., Intxaurrondo, A., Rodríguez, H.,
  Lopez~Martin, J.A., Villegas, M., Krallinger, M.: Automatic de-identification
  of medical texts in spanish: the meddocan track, corpus, guidelines, methods
  and evaluation of results. In: Proceedings of the Iberian Languages
  Evaluation Forum (IberLEF 2019). vol.~TBA, p.~TBA. CEUR Workshop Proceedings
  (CEUR-WS.org), Bilbao, Spain (Sep 2019)

\bibitem{elmo/Peters18}
Peters, M.E., Neumann, M., Iyyer, M., Gardner, M., Clark, C., Lee, K.,
  Zettlemoyer, L.: Deep contextualized word representations. In: Proc. of NAACL
  (2018)

\bibitem{emb/Soares19}
Soares, F., Villegas, M., Gonzalez-Agirre, A., Krallinger, M.,
  Armengol-Estap{\'e}, J.: Medical word embeddings for {S}panish: Development
  and evaluation (Jun 2019), \url{https://www.aclweb.org/anthology/W19-1916}

\bibitem{i2b2/Stubs15}
Stubbs, A., Uzuner, {\"O}.: Annotating longitudinal clinical narratives for
  de-identification: The 2014 i2b2/uthealth corpus. Journal of biomedical
  informatics  \textbf{58},  S20--S29 (2015)

\bibitem{ib2b/Yang14}
Yang, H., Garibaldi, J.M.: Automatic detection of protected health information
  from clinic narratives. Journal of Biomedical Informatics  \textbf{58},  S30
  -- S38 (2015). \doi{https://doi.org/10.1016/j.jbi.2015.06.015}

\end{thebibliography}
\end{document}